\documentclass[10pt,twocolumn,letterpaper]{article}

\usepackage{iccv}
\usepackage{times}
\usepackage{epsfig}
\usepackage{graphicx}
\usepackage{amsmath}
\usepackage{amssymb}
\usepackage[utf8]{inputenc}
\usepackage{algorithm}
\usepackage{algpseudocode}

\usepackage{makecell}
\usepackage{xcolor}
\usepackage{amsfonts}
\usepackage{multirow}
\usepackage{float}
\usepackage{enumitem}
\usepackage{comment}
\usepackage[accsupp]{axessibility} 

\usepackage{bbding}
\usepackage{xcolor}

\usepackage{utfsym}
\usepackage{pifont}
\newcommand*\rot{\rotatebox{90}}

\usepackage{lipsum}
\newcommand\blfootnote[1]{%
  \begingroup
  \renewcommand\thefootnote{}\footnote{#1}%
  \addtocounter{footnote}{-1}%
  \endgroup
}

\usepackage[breaklinks=true,bookmarks=false,hyperfootnotes=false]{hyperref}

\iccvfinalcopy 

\def\iccvPaperID{****} 

\ificcvfinal\fi

\usepackage{pdfpages} 
\usepackage{pgffor} 

\makeatletter
\AtBeginDocument{\let\LS@rot\@undefined}
\makeatother

\begin{document}

\title{Hard No-Box Adversarial Attack on Skeleton-Based Human Action Recognition with Skeleton-Motion-Informed Gradient}

\author{Zhengzhi Lu$^{1,2~\dag}$ 
\quad
He Wang$^3$
\quad
Ziyi Chang$^{1}$
\quad
Guoan Yang$^2$
\quad
Hubert P. H. Shum$^{1~\ddag}$\\
$^1$Durham University, UK \quad $^2$Xi’an Jiaotong University, China \quad $^3$University College London, UK\\
{\tt\small lu947867114@stu.xjtu.edu.cn} \quad {\tt\small he\underline{~}wang@ucl.ac.uk} \quad {\tt\small ziyi.chang@durham.ac.uk}  \\{\tt\small gayang@mail.xjtu.edu.cn} \quad {\tt\small hubert.shum@durham.ac.uk}  
}

\maketitle
\ificcvfinal\thispagestyle{empty}\fi

\begin{abstract}
   Recently, methods for skeleton-based human activity recognition have been shown to be vulnerable to adversarial attacks. However, these attack methods require either the full knowledge of the victim (i.e. white-box attacks), access to training data (i.e. transfer-based attacks) or frequent model queries (i.e. black-box attacks). All their requirements are highly restrictive, raising the question of how detrimental the vulnerability is. In this paper, we show that the vulnerability indeed exists. To this end, we consider a new attack task: the attacker has no access to the victim model or the training data or labels, where we coin the term hard no-box attack. Specifically, we first learn a motion manifold where we define an adversarial loss to compute a new gradient for the attack, named skeleton-motion-informed (SMI) gradient. Our gradient contains information of the motion dynamics, which is different from existing gradient-based attack methods that compute the loss gradient assuming each dimension in the data is independent. The SMI gradient can augment many gradient-based attack methods, leading to a new family of no-box attack methods. Extensive evaluation and comparison show that our method imposes a real threat to existing classifiers. They also show that the SMI gradient improves the transferability and imperceptibility of adversarial samples in both no-box and transfer-based black-box settings.\vspace{-1em}
\end{abstract}

\section{Introduction}
\blfootnote{$^\dag$ This work was conducted during the visit to the Durham University.}
\blfootnote{$^\ddag$ Corresponding Author}Deep learning models are vulnerable to adversarial attacks, which compute data perturbations strategically to fool trained networks. Since its discovery~\cite{szegedy2013intriguing}, a wide variety of models in different tasks have been attacked~\cite{RN9}, raising severe concerns as these perturbations are imperceptible to humans. Recently, the adversarial attack in skeleton-based human activity recognition (S-HAR) has attracted attention as skeletal data have been widely used in security-critical applications such as sports analysis, bio-mechanics, surveillance, and human-computer interactions \cite{ren2020survey}.

\begin{table}[]
\setlength\tabcolsep{4pt}
\footnotesize
\begin{tabular}{cccccc}
\Xhline{1.5pt}
\begin{tabular}[c]{@{}c@{}}Information\\ Accessible\end{tabular} &
  \begin{tabular}[c]{@{}c@{}}White-\\Box \end{tabular} &
  \begin{tabular}[c]{@{}c@{}}Queried\\ Black-Box\end{tabular} &
  \begin{tabular}[c]{@{}c@{}}Transferred\\ Black-Box\end{tabular} &
  \begin{tabular}[c]{@{}c@{}}No-\\Box \end{tabular} &
  \begin{tabular}[c]{@{}c@{}}\textbf{Hard} \\\textbf{No-Box}\end{tabular} \\ \hline
Model  Parameters   & \checkmark & $\times$ & $\times$ & $\times$ & $\times$ \\
Queries of  Victims & \checkmark & \checkmark & $\times$ & $\times$ & $\times$ \\
Training Samples  & $\times$ & $\times$ & \checkmark & $\times$ & $\times$ \\
Labels            &\checkmark  & \checkmark & \checkmark & \checkmark & $\times$\\\Xhline{1.5pt}
\end{tabular}
\caption{Comparisons on different settings of adversarial attacks. $\checkmark$ and $\times$ indicate if a method needs to access the corresponding information.}\vspace{-1em}
\label{compare}
\end{table}

Existing attacks in S-HAR are categorized into white-box and black-box approaches. White-box approaches require prior knowledge of the full details of a victim model~\cite{RN10, RN5} while black-box approaches require a large number of queries to the victim model~\cite{RN7} or the access to training data and labels~\cite{RN5}. On the one hand, the victim model details and the training data and labels are unlikely to be available to the attacker in real-world scenarios. On the other hand, making frequent and numerous queries (e.g. tens of thousands) to the victim model is time-consuming and raises suspicion. In other words, the settings of existing S-HAR attacks are overly restrictive. A key to a successful attack is to reduce the required information of the victim model, training data and labels.  

In this paper, we introduce a new threat model that requires no access to the victim model, training data or labels. We name the new threat model the \textit{hard no-box attack}, differentiating from the recent no-box attack on images \cite{RN13} that does not require access to the victim model but still needs access to the labels (i.e. soft no-box attack). Table \ref{compare} demonstrates the comparison on different settings of adversarial attacks. Among all attack settings, our hard no-box attack requires the least amount of knowledge, as it can only access the testing data without labels. Designing such an attack is nontrivial and challenging. Without access to the victim model, the attack method cannot rely on the gradient of a classification loss \cite{RN4}, data manipulation during training \cite{saha2020hidden}, and the feedback of a classifier \cite{brendel2018decisionbased}. The challenge is further exacerbated by the requirement of no label and training sample access, where no surrogate model can be trained to attack or estimate the data distribution.

To tackle the challenges, we propose a contrastive learning (CL) ~\cite{RN15} solution with a manifold-based no-box adversarial loss. First, we introduce a new application of CL to learn a latent data manifold where similar samples are naturally aggregated while dissimilar samples are dispersed without the need of class labels. It provides a good description of sample similarity that facilitates generating skeletal adversarial samples. CL is suitable for hard no-box attack settings due to its ability to capture the discriminative high-level features under our restricted attack conditions. Second, we compute the perturbation to drag a data sample away from its similar neighbors in the latent space, bounded by a pre-defined budget. In particular, we design a new no-box adversarial loss to maximize each adversary's dissimilarity with positive samples while minimizing its similarity with negative samples. The loss serves as guidance for the adversary search in our gradient-based attack scheme.

While gradient-based attack methods like I-FGSM~\cite{RN11} are shown to be effective on S-HAR attacks~\cite{RN5, RN10}, the gradient is computed based on the victim model and the labels, making it unsuitable for hard no-box attacks. Since adversarial samples are likely to lie in or near the motion manifold~\cite{RN7}, ideally, we want to explore along the manifold. That is, the computation of adversarial loss gradient should consider the local motion manifold.

To this end, we propose to explicitly model motion dynamics for describing the local manifold around a given motion. Specifically, we introduce the skeleton-motion-informed (SMI) gradient that employs dynamics models (e.g. Markovian and autoregressive) to represent motion dynamics for the loss gradient computation. As a result, while existing methods generally assume each dimension in a data sample to be independent when computing the loss gradient, SMI gradient explicitly considers the dependency between frames in time. Furthermore, the SMI gradient is compatible with existing gradient-based methods including I-FGSM and MI-FGSM~\cite{RN20}, allowing us to effectively construct a new family of no-box attack methods.

Extensive experiments show that our method generates effective adversarial samples that successfully attack various victim models across datasets (HDM05, NTU60 and NTU120). Our SMI-gradient based attacks improve the attack transferability in both no-box and transferred black-box settings, with better imperceptibility. Codes are available in {\color{blue} \url{https://github.com/luyg45/HardNoBoxAttack}} and our contributions are:

\begin{itemize}
[noitemsep,topsep=0pt,labelindent=0cm,leftmargin=0.4cm]
    \item We confirm the S-HAR threat by introducing a new hard no-box attack and proposing the first method to generate adversarial samples without access to the victim model or training data or labels, to the best of our knowledge. 
    \item We propose a new skeleton-motion-informed gradient that guides the adversary search along the motion manifold, explicitly considering the spatial-temporal nature of skeletal motions.
    \item We present a family of novel gradient-based attack strategies facilitated by the new gradient, improving the transferability and imperceptibility of adversarial samples in no-box and transferred black-box attacks.
\end{itemize}
                                  
\section{Related Works}
\textbf{Skeleton-Based Human Action Recognition}
S-HAR has attracted considerable attention in many applications \cite{ren2020survey} where deep learning-based approaches have achieved state-of-the-art performance \cite{soo2017interpretable, ke2017new}.
Recurrent neural networks are employed to model the temporal domain of human motions~\cite{du2015hierarchical, RN34}. Furthermore, unlike images and videos, the skeleton has a graph structure, so graph convolutional networks have shown to be effective in modelling the spatial or spatial-temporal features \cite{RN33, RN28}. The effectiveness is generally achieved by considering the skeleton as a topological graph where the joints and bones correspond to nodes and edges \cite{men21quadruple}. Improved graph designs and network architectures are subsequently proposed \cite{RN39, RN37,zhou2023self,zhang2022hierarchical}.

\textbf{Adversarial Attacks on Skeletons}
Adversarial attacks were initially introduced in \cite{szegedy2013intriguing}, which showcases the vulnerability of deep neural networks and has been extended to other data types. Generally, adversarial attack is a special technique of data augmentation that aims to reveal the vulnerability of a system by finding new samples, while other data augmentation techniques may have diverse purposes, e.g. training efficiency and inference performance \cite{Shanmugam_2021_ICCV}.
Recently, the attack on S-HAR has received increasing attention. Wang et al. \cite{RN5} analyzed the perceptibility of adversarial skeletal samples and proposed a new perceptual loss. Liu et al. \cite{RN10} focused on GCN-based models and utilized generative adversarial networks to synthesize adversarial examples. Tanaka et al. \cite{RN6} proposed a new lower-dimensional attack, in which only the length of bones could be perturbed. These methods all require the complete knowledge of victim models, a setting known as the white-box attack. In contrast, Diao et.al \cite{RN7} introduced the first black-box S-HAR attack method, which searches motion manifolds for adversaries. Still, black-box attacks need to frequently query the victim models, which can be infeasible in real-world systems. In contrast, we consider a more practical threat setting named the hard no-box attack, where an attacker only has access to unlabeled skeletal data.

\textbf{Gradient-Based Attack Strategies}
The core component of adversarial attacks is to generate adversarial samples \cite{RN9}. Gradient-based attack methods have been widely used to introduce perturbations to a sample following the direction of the loss gradient. Goodfellow et al. \cite{RN4} proposed the fast gradient sign method (FGSM) that perturbs a sample by a single step along the loss gradient. Kurakin et al. \cite{RN11} proposed  I-FGSM by extending FGSM to an iterative process. Dong et al. \cite{RN20} presented MI-FGSM by adding momentum to the gradient, which boosted the transferability of adversarial samples. Xie et al. \cite{xie2019improving} applied diversified augmentations to the inputs before each iteration to craft more transferable samples. While these gradient-based strategies are successful in static data and have been adapted to skeletal motions, they neglect the dependency between frames for gradient computation, 
which is crucial in time series. This motivates us to propose our skeleton-motion-informed attack strategies, which explicitly model the motion dynamics in the temporal domain~\cite{RN25,wang_defending_2022}.

\section{Hard No-Box Attack for Skeletal Data}
\begin{figure}[tb] 
\footnotesize
\centering
\includegraphics[width=0.48\textwidth]{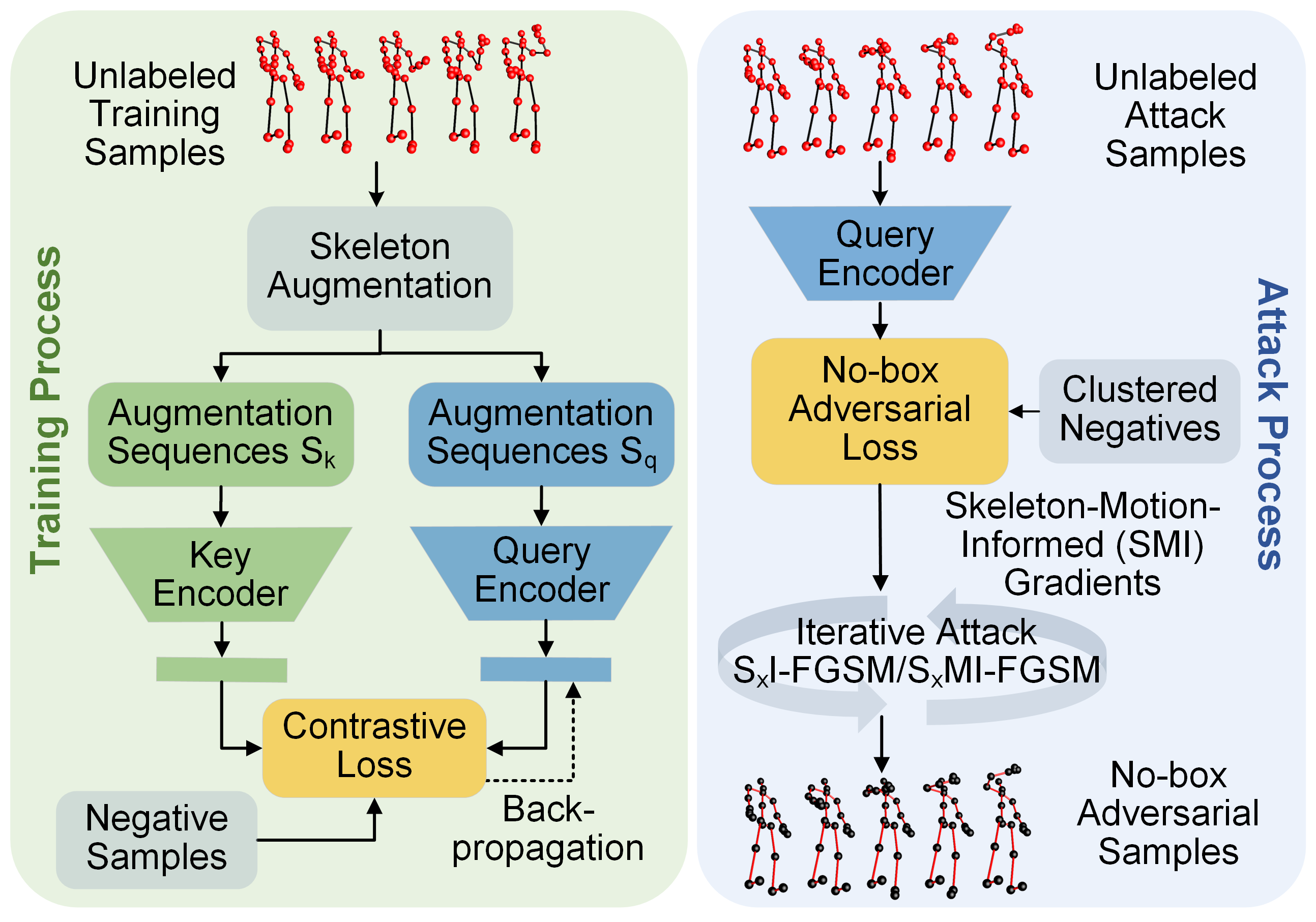} 
\caption{The training (left) and attack (right) processes for the hard no-box attack. The trained query encoder in the training process is used for attacks in the attack process.}\vspace{-1em} 
\label{Skecon} 
\end{figure}

Figure \ref{Skecon} shows the overview of our method. The left part is the training process where we adopt contrastive learning to obtain a latent data manifold to distinguish data samples. The attack process is shown on the right-hand side. We first design a new no-box adversarial loss in the trained latent space to guide the adversary search using samples that are dissimilar to the attacked sample. Then we propose a novel skeleton-motion-informed gradient and a new family of attack methods for generating adversarial samples.

\subsection{Contrastive Learning for Motion Manifold} 

While the fundamental idea of adversarial attacks is to perturb a data sample to cross class boundaries, such boundaries cannot be estimated for hard no-box attacks due to the lack of labels. To estimate such boundaries without labels, we present a new application of contrastive learning (CL) \cite{RN15} to aggregate similar data samples as soft class boundaries in latent space. Such boundaries enable us to adversarially perturb a sample to cross boundaries. We also train an encoder to extract discriminative high-level features for the motion manifold in latent space. Overall, our CL constructs boundaries in latent space without labels via aggregating similar samples and segregating dissimilar samples. Our attack is guided by the dissimilarity of high-level features between samples for generating adversarial samples, instead of using class boundaries to lead the attack \cite{brendel2018decisionbased}.

To incorporate both spatial and temporal information, we train an encoder (Fig.~\ref{Skecon} left) based on adaptive graph convolutional network (AGCN) ~\cite{RN33}. To force encoders to focus on high-level features, we apply skeleton-specific data augmentations to an input sequence $S$ and obtain two different views $S_q$ and $S_k$. Augmentations include spatial operations (e.g. pose transformations, joint jittering, etc.) and temporal operations (e.g. temporal crop and resize) \cite{RN15} (detailed in supplementary material). Then, we feed $S_q$ and $S_k$ into the query encoder $f_q$ and the key encoder $f_k$ respectively for the info-noise-contrastive estimation (InfoNCE) \cite{RN32}:
\begin{flalign}
&L_{contrast}=\\
&-\log \frac{\exp \left(f_{q}(S_q) \cdot f_{k}(S_k) / \tau\right)}{\exp \left(f_{q}(S_q) \cdot f_{k}(S_k) / \tau\right)+\sum\limits_{F_{n} \sim N} \exp \left(f_{q}(S_q) \cdot F_{n} / \tau\right)} , \nonumber
\end{flalign}
where $\tau$ is the temperature parameter, $N$ is the dynamic queue storing the features of negative samples $F_{n}$ obtained in the training process. After training, we use the query encoder $f_{q}$, which encodes the motion manifold, for attack.

\subsection{Adversarial Loss for Unlabeled Skeletal Data}
The adversarial loss of hard no-box attack is significantly different from most existing methods that heavily depend on labels and class boundaries. Since class labels and class boundaries are unavailable in hard no-box attacks, we utilize data samples that are dissimilar to a given sample (i.e. \textit{negative samples}) for defining the adversarial loss. Correspondingly, samples that are similar to the given sample are considered as \textit{positive samples}. We argue when a given sample is perturbed towards its negative samples and away from its positive samples, it tends to become an adversary. This is because the negative samples generally indicate the high-density areas of other classes in the latent space. 

The hard no-box adversarial loss is designed as: 
\begin{equation}
\begin{split}
&L_{adv}=-\log \frac{\exp \left[Sim\left(f_{q}\left(s\right), f_{q}\left(\tilde{s}\right)\right)\right]}{\sum_{j} \exp \left[Sim\left(f_{q}\left(s\right), f_{q}\left(\tilde{s}_{j}\right)\right)\right]}, 
\end{split}
\label{advloss}
\end{equation}
where $Sim$ is the cosine similarity, $s$ is the adversarial sample to be computed, $\widetilde{s}$ is the clean sample regarded as the positive sample, and $\widetilde{s_j}$ are the negative samples. Maximizing $L_{adv}$ moves $s$ away from $\widetilde{s}$ and towards $\widetilde{s_j}$ in the latent manifold. With $L_{adv}$, gradient-based attacks 
are employed.


To maximize Eq.~\ref{advloss}, the selection of negative samples $\widetilde{s_j}$ is crucial and we design a method tailored for no-box attacks. Existing work \cite{fan2021does} utilizes cluster-fit \cite{yan2020clusterfit} to generate pseudo labels for selecting negative samples during adversarial training, which is less suitable for the no-box attack as it requires another pretrained off-line encoder to obtain pseudo labels. Instead, we adapt K-means, an unsupervised method, to select negatives, removing the need of any pre-training. We discard $Q$ clusters whose cluster centers are the closest to the input sample, mitigating the risk of misleading attacks. The remaining cluster centers are considered as the negative samples $\tilde{s}_j$.

\section{Skeleton-Motion-Informed Gradient}
 Existing gradient-based attack methods treat each dimension of the data as an independent variable, i.e. raw gradient. Attacks based on raw gradients tend to drag a sample away from the data manifold \cite{feinman2017detecting}. With the guidance of class boundaries and a limit on the perturbation budget, the raw gradient can still find deceiving adversaries. However, this setting is infeasible in hard no-box attacks. Without class boundaries, raw gradients that point to the negative samples can drag the adversary far away from the manifold. This is because while the perturbations are towards negative samples, they are not necessarily in a direction orthogonal to the class boundary. Consequently, larger perturbations are needed to cross the boundary, leading to adversaries being far off the manifold. This creates the need to constrain the perturbation within or near the manifold, at least locally. Since the motion manifold is constrained by the motion dynamics \cite{RN12}, we argue that the gradient needs to explicitly capture the dynamics. Therefore, we propose a new gradient named skeleton-motion-informed (SMI) gradient, capturing the manifold information that has been largely ignored by existing methods in loss gradient computation.

\subsection{Dynamics in the Gradient Structure}

Given a skeletal sequence $S=[S_{1}, S_{2}, \cdots, S_{t}]$ and the adversarial loss $J(S)$, a straightforward but effective strategy to craft adversarial perturbations is the gradient-based attack \cite{RN9}. It utilizes backpropagation of the loss function $\nabla J(S)$ to iteratively change input samples $S$:

\begin{equation}
\hat{S}=S+\alpha \cdot \operatorname{sign}\left(\nabla J(S)\right) ,
\label{attack}
\end{equation}
where $\alpha$ is the attack step size and $\hat{S}$ is the adversarial samples. In skeletal motions, this attack gradient $\nabla J(S)$ consists of a set of partial derivatives over all frames: 
\begin{equation}
\nabla J(S)=\left[\frac{\partial J(S)}{\partial S_{1}}, \frac{\partial J(S)}{\partial S_{2}}, \cdots, \frac{\partial J(S)}{\partial S_{t}}\right]. 
\label{partial derivatives}
\end{equation} 
The partial derivative $\frac{\partial J(S)}{\partial S_{t}}$ assumes each frame is independent, and this is the raw gradient employed in existing methods~\cite{RN5}. However, human motions contain rich dynamics so that the system can be described as $S_{t} = f(S_{<t})$. So far, various dynamics models have been attempted to model human motions, such as Markovian models~\cite{tang_realtime_2022,wang_energy_2015}, autoregressive models~\cite{RN25}, and many-to-many mapping~\cite{RN12}, all of which can capture the dynamics at different scales in time. 
We explore these models to reveal the missing dynamics in the structure of the raw gradient and propose our SMI-gradients that consider motion dynamics.

\paragraph{Markovian Model} We assume the motion dynamics can be captured by a Markovian model, i.e. $S_t = f_{d1}(S_{t-1})$. This allows us to derive the 1st-order SMI-gradient: 
\begin{equation}
\left(\frac{\partial J(S)}{\partial S_{t-1}}\right)_{d1}=\frac{\partial J(S)}{\partial S_{t-1}}+\frac{\partial J(S)}{\partial S_{t}} \cdot \frac{d S_{t}}{d S_{t-1}},
\label{first-order real gradient}
\end{equation}
where $\frac{d S_{t}}{d S_{t-1}}$ is the temporal relationship between two consecutive frames that will be instantiated. Eq. \ref{first-order real gradient} shows that the attack gradient along the motion manifold needs to consider the first-order information in the motion, e.g. velocity.
 
\paragraph{Autoregressive Model} 
Besides the first-order dynamics, we also model the second-order dynamics by assuming $S_t = f_{d2}(S_{t-1}, S_{t-2})$, as $2nd$-order dynamics (i.e. joint acceleration) capture the smooth temporal dynamics of skeletal motion \cite{RN5}. We extend Eq. \ref{first-order real gradient} as: 

\begin{equation}
\begin{split}
\left(\frac{\partial J(S)}{\partial S_{t-2}}\right)_{d2}=\frac{\partial J(S)}{\partial S_{t-2}}+\frac{\partial J(S)}{\partial S_{t-1}} \cdot \frac{\partial S_{t-1}}{\partial S_{t-2}}\\+\frac{\partial J(S)}{\partial S_{t}} \cdot (\frac{\partial S_{t}}{\partial S_{t-2}}+\frac{\partial S_{t}}{\partial S_{t-1}}\cdot\frac{\partial S_{t-1}}{\partial S_{t-2}}).
\label{second-order real gradient}
\end{split}
\end{equation}

While higher-order models can also be considered, there is an empirical evidence that the first three orders are the most important in skeletal motion adversarial attack~\cite{RN5}. Therefore, we express the SMI gradients of the whole skeletal sequence as $(\nabla J(S))_{d1}=\left[ (\frac{\partial J(S)}{\partial S_{1}})_{d1}, (\frac{\partial J(S)}{\partial S_{2}})_{d1}, \cdots, (\frac{\partial J(S)}{\partial S_{t}})_{d1}\right]$ and $(\nabla J(S))_{d2}=\left[ (\frac{\partial J(S)}{\partial S_{1}})_{d2}, (\frac{\partial J(S)}{\partial S_{2}})_{d2}, \cdots, (\frac{\partial J(S)}{\partial S_{t}})_{d2}\right]$.

\subsection{Time-Varying Autoregressive Models for Dynamics}
We employ explicit models~\cite{RN25} to compute the dynamics-related derivatives in SMI gradients. While implicit models~\cite{tang_realtime_2022} may also be considered, they would require another network to be trained, making them less preferable in hard no-box attacks. 
To realize $f_{d1}$ and $f_{d2}$, we use time-varying autoregressive models (TV-AR) \cite{bringmann2017changing}, which effectively estimates the dynamics of skeleton sequence ~\cite{RN25} due to its ability of modelling the temporal non-stationary signals: 
\begin{equation}
f_{d1}: \text{  }S_{t}=A_{t}\cdot S_{t-1}+ B_{t} + \gamma_{t}, 
\label{AR1}
\end{equation}
\begin{equation}
f_{d2}: \text{  }S_{t}=C_{t}\cdot S_{t-1}+ D_{t}\cdot S_{t-2} +E_{t}+ \gamma_{t} ,
\label{AR2}
\end{equation}
where Eq. \ref{AR1} and Eq. \ref{AR2} are denoted as TV-AR(1) and TV-AR(2) respectively. The model parameters $\beta_{t}^{1}=[A_{t},B_{t}]$ and $\beta_{t}^{2}=[C_{t},D_{t},E_{t}]$ are all time-varying parameters and determined by data-fitting. $\gamma_{t}$ is a time-dependent white noise representing the dynamics of stochasticity.

Using Eq. \ref{AR1} to compute $\frac{\partial S_{t}}{\partial S_{t-1}}$, Eq. \ref{first-order real gradient} becomes:
\begin{equation}
\left(\frac{\partial J(S)}{\partial S_{t-1}}\right)_{d1}=\frac{\partial J(S)}{\partial S_{t-1}}+\frac{\partial J(S)}{\partial S_{t}} \cdot {A_{t}}.
\label{first-order AR}
\end{equation}
Similarly, using Eq. \ref{AR2}, we can compute $C_{t}=\frac{\partial S_{t}}{\partial S_{t-1}}$ and $D_{t}=\frac{\partial S_{t}}{\partial S_{t-2}}$. For $\frac{\partial S_{t-1}}{\partial S_{t-2}}$, we use $S_{t-1}={C_{t-1}}\cdot S_{t-2}+ {D_{t-1}}\cdot S_{t-3} +{E_{t-1}}+ \gamma_t$ to compute it: ${C_{t-1}}=\frac{\partial S_{t-1}}{\partial S_{t-2}}$. Then, Eq. \ref{second-order real gradient} becomes: 
\begin{equation}
\footnotesize
\left(\frac{\partial J(S)}{\partial S_{t-2}}\right)_{d2}=\frac{\partial J(S)}{\partial S_{t-2}}+\frac{\partial J(S)}{\partial S_{t-1}} \cdot {C_{t-1}}+\frac{\partial J(S)}{\partial S_{t}} \cdot (D_{t}+C_{t}\cdot {C_{t-1}}) .
\label{second-order AR}
\end{equation}

\section{Skeleton-Motion-Informed Attack}
We construct new gradient-based attack methods based on our novel SMI gradient. Due to its compatibility, our proposed gradient can be integrated with most existing gradient-based methods. We select I-FGSM and MI-FGSM, which have been proven for their efficiency on S-HAR attacks \cite{RN5, RN10}. We augment them with first and second-order SMI-gradients, leading to four new attack methods. 

\textbf{Fast Gradient Sign Methods (FGSM)} FGSM \cite{RN4} is a single-step attack method that generates the adversarial samples $\hat{S}=S+r$ by maximizing the adversarial loss function $J(S)$, where $r$ denotes an adversarial perturbation that is constrained within a budget $\lVert r\rVert_p<\epsilon$, where $\lVert \cdot\rVert_p$ denotes the $l_p$-norm. One variant of FGSM is the Iterative Fast Gradient Sign Method (I-FGSM) \cite{RN11}, which extends FGSM to an iterative process:
\begin{equation}
\hat{S}^{i+1}=\hat{S^{i}}+\alpha \cdot \operatorname{sign}\left(\nabla_{s} J(S)\right) ,
\end{equation}
where $\alpha$ is the attack step size and $i$ means iteration. Another variant is MI-FGSM \cite{RN20}, which considers the momentum of the attack to avoid local maxima:
\begin{equation}
g^{i+1}=\mu \cdot g^{i}+\frac{\nabla_{s}J\left(S\right)}{\left\|\nabla_{s} J\left(S\right)\right\|_{1}}
\end{equation}
\begin{equation}
\hat{S}^{i+1}=\hat{S^{i}}+\alpha \cdot \operatorname{sign}\left(g^{i+1}\right) ,
\end{equation}
where $\mu$ is the momentum decay factor and $g_{i}$ is the gradient in iteration $i$.

\begin{algorithm}[htb]
\caption{$\rm S_{1}$I-FGSM and $\rm S_{2}$I-FGSM}\label{algorithm}
\begin{algorithmic}[1]
\footnotesize
\Require An encoder $k$ with a loss function $J$; a skeletal sequence sample $S$; the size of attack step $\alpha$; iterations $I$; the budget of perturbation $\epsilon$.
\Ensure An adversarial example $\hat{S}$ with $\lVert \hat{S}-S\rVert_p<\epsilon$.
\State Initialization: $\hat{S}^{0}=S$;
\State Fitting $S$ with TV-AR model to obtain the time-varying parameters $\beta_{t}$;
\For{$i=0$ to $I-1$}
    \State Input $\hat{S}^{i}$ to $k$;  
    \State Obtain the raw gradient $\nabla J(\hat{S}^{i})$  on $J$;
    \State Calculate the SMI gradient $(\nabla J(\hat{S}^{i}))_{d1}$ with Eq. \ref{first-order AR}, or \mbox{\quad} $(\nabla J(\hat{S}^{i}))_{d2}$ with Eq. \ref{second-order AR}, using $\beta_{t}$ and $\nabla J(\hat{S}^{i})$;
    \State Update $\hat{S}^{i+1}$ by applying the sign gradient as:
    \begin{equation}
    \begin{split}
         &\hat{S}^{i+1}=\hat{S}^{i}+\alpha \cdot \operatorname{sign}\left(\nabla J(\hat{S}^{i}))_{d1}\right), or\\
   &\hat{S}^{i+1}=\hat{S}^{i}+\alpha \cdot \operatorname{sign}\left(\nabla J(\hat{S}^{i}))_{d2}\right).   
    \end{split}
    \end{equation}
\EndFor
\State \textbf{return} $\hat{S}=\hat{S}^{I}$
\end{algorithmic}
\end{algorithm}

\textbf{SMI-gradient Based Attacks}
We replace the original gradient $\nabla J(S)$ in I-FGSM and MI-FGSM with our SMI gradient $(\nabla J(S))_{d1}$ or $(\nabla J(S))_{d2}$. This creates four new dynamic attack strategies: first-order SMI I-FGSM ($\rm S_{1}$I-FGSM), second-order SMI I-FGSM ($\rm S_{2}$I-FGSM), first-order SMI MI-FGSM ($\rm S_{1}$MI-FGSM), and second-order SMI MI-FGSM ($\rm S_{2}$MI-FGSM). The processes of SI-FGSM are shown in Algorithm~\ref{algorithm}. The algorithm of SMI-FGSM can be found in the supplementary material. 


\begin{table*}[htb]
\setlength\tabcolsep{4pt}
\scriptsize
\centering
\begin{tabular}{|c|c|cc|cccccc|}
\cline{2-10}
  \multicolumn{1}{c|}{} &
  \makecell{Victim\\models} &
  \makecell{Self-sup\\Attacker} &
  \makecell{AGCN\\Attacker} &
  \makecell{No-box\\I-FGSM} &
  \makecell{No-box \\$\rm S_{1}$I-FGSM} &
  \makecell{No-box \\$\rm S_{2}$I-FGSM} &
  \makecell{No-box\\MI-FGSM} &
  \makecell{No-box \\$\rm S_{1}$MI-FGSM} &
  \makecell{No-box \\$\rm S_{2}$MI-FGSM} \\ \hline
\multirow{6}{*}{\rot{$\epsilon = 0.01$}} & js-AGCN & 11.21\% & ——      & 26.05\% & 28.09\% & 30.87\% & 30.68\% & 34.75\% & \textbf{36.58}\% \\
& 2s-AGCN & 5.36\%  & ——      & 13.94\% & 15.04\% & 16.45\% & 16.47\% & 18.23\% & \textbf{19.31}\% \\
& ST-GCN  & 3.57\%  & 12.93\% & 9.55\%  & 9.86\%  & 9.96\%  & 11.11\%  & 11.36\% & \textbf{11.56}\% \\
& MS-G3D  & 8.39\%  & 35.23\% & 10.57\% & 11.14\% & 11.69\% & 11.76\% & 12.85\% & \textbf{14.11}\% \\
& SGN     & 21.25\% & 26.03\%      & 34.09\% & 34.46\% & 35.23\% & 38.49\% & 38.75\% & \textbf{38.81}\% \\
& ASGCN   & 5.69\%  & 20.87\%      & 13.92\% & 14.85\% & 14.67\% & 15.95\% & 16.82\% & \textbf{17.75}\% \\ 
\hline\hline
\multirow{6}{*}{\rot{$\epsilon = 0.008$}} &  js-AGCN & 10.12\% & ——      & 22.84\% & 24.36\% & 26.46\% & 25.70\% & 29.64\% & \textbf{30.88}\% \\
& 2s-AGCN & 5.04\%  & ——      & 11.36\% & 12.19\% & 13.03\% & 12.30\% & 13.85\% & \textbf{15.56}\% \\
& ST-GCN  & 3.26\%  & 10.33\% & 7.87\%  & 7.99\%  & 8.04\%  & 8.84\%  & 9.07\% & \textbf{9.19}\%  \\
& MS-G3D  & 5.19\%  & 31.28\% & 9.18\% & 9.51\% & 9.99\% & 10.01\% & 10.26\% & \textbf{12.98}\% \\
& SGN     & 20.95\%      & 23.30\%      & 29.61\% & 30.20\% & 30.74\% & 33.32\% & 33.63\% & \textbf{34.54}\% \\
& ASGCN   & 5.54\%      & 18.77\%      & 11.42\% & 12.05\% & 12.29\% & 12.77\% & 13.69\% & \textbf{14.37}\% \\ 
\hline\hline
\multirow{6}{*}{\rot{$\epsilon = 0.006$}} & js-AGCN & 7.19\% & ——      & 19.70\% & 20.32\% & 21.23\% & 20.04\% & 22.76\% & \textbf{24.33}\% \\
& 2s-AGCN & 3.70\% & ——      & 7.93\%  & 9.80\%  & 10.56\% & 9.80\%  & 11.26\% & \textbf{12.66}\% \\
& ST-GCN  & 2.46\% & 7.88\%  & 5.65\%  & 5.85\%  & 5.97\%  & 6.25\%  & 6.54\% & \textbf{6.66}\%  \\
& MS-G3D  & 4.46\% & 23.15\% & 7.30\% & 7.84\% & 7.61\% & 7.94\% & 8.05\% & \textbf{8.39}\% \\
& SGN     & 16.76\% & 19.93\% & 23.92\% & 24.57\% & 25.75\% & 26.74\% & 27.04\% & \textbf{27.64}\% \\
& ASGCN   & 3.94\% & 14.30\% & 8.79\%  & 9.23\% & 9.07\%  & 9.71\% & 10.29\% & \textbf{10.90}\% \\ 
\hline
\end{tabular}
\caption{The fooling rate of different methods on the target models in NTU60, where $\epsilon$ is the perturbation budget.}\vspace{-1em}
\label{1}
\end{table*}

\begin{table*}[htb]
\setlength\tabcolsep{4pt}
\scriptsize
\centering
\begin{tabular}{|c|c|cc|cccccc|}
\cline{2-10}
  \multicolumn{1}{c|}{} &
  \makecell{Victim\\models} &
  \makecell{Self-sup\\Attacker} &
  \makecell{AGCN\\Attacker} &
  \makecell{No-box\\I-FGSM} &
  \makecell{No-box \\$\rm S_{1}$I-FGSM} &
  \makecell{No-box \\$\rm S_{2}$I-FGSM} &
  \makecell{No-box\\MI-FGSM} &
  \makecell{No-box \\$\rm S_{1}$MI-FGSM} &
  \makecell{No-box \\$\rm S_{2}$MI-FGSM} \\ \hline
\multirow{6}{*}{\rot{$\epsilon = 0.01$}} & js-AGCN & 9.97\% & ——      & 23.92\% & 24.64\% & 25.79\% & 27.26\% & 27.93\% & \textbf{29.07}\% \\
& 2s-AGCN & 6.38\%  & ——      & 20.09\% & 21.11\% & 22.06\% & 23.51\% & 24.63\% & \textbf{24.96}\% \\
& ST-GCN  & 12.18\%  & 23.84\% & 23.53\%  & 24.73\%  & 25.77\%  & 26.95\%  & 27.31\% & \textbf{28.76}\% \\
& MS-G3D  & 10.63\%  & 24.63\% & 19.07\% & 20.11\% & 20.20\% & 21.57\% & 21.95\% & \textbf{22.73}\% \\
& SGN     & 19.65\% & 37.90\%      & 31.43\% & 32.64\% & 33.75\% & 38.47\% & 37.96\% & \textbf{38.85}\% \\
& ASGCN   & 7.29\%  & 24.15\%      & 18.03\% & 19.29\% & 20.37\% & 19.88\% & 20.22\% & \textbf{21.60}\% \\ 
\hline\hline
\multirow{6}{*}{\rot{$\epsilon = 0.008$}} &  js-AGCN & 9.04\% & ——      & 21.58\% & 21.77\% & 22.47\% & 24.28\% & 24.62\% & \textbf{25.74}\% \\
& 2s-AGCN & 5.79\%  & ——      & 15.71\% & 15.23\% & 15.94\% & 18.68\% & 18.76\% & \textbf{19.57}\% \\
& ST-GCN  & 11.23\%  & 21.35\% & 20.74\%  & 21.51\%  & 22.22\%  & 23.52\%  & 24.65\% & \textbf{24.87}\%  \\
& MS-G3D  & 9.98\%  & 21.73\% & 17.06\% & 17.74\% & 17.92\% & 19.53\% & 19.29\% & \textbf{20.33}\% \\
& SGN     & 17.59\%      & 35.88\%      & 27.38\% & 28.17\% & 28.65\% & 32.43\% & \textbf{33.06}\% & 32.55\% \\
& ASGCN   & 6.87\%      & 21.87\%      & 15.73\% & 16.93\% & 17.62\% & 17.41\% & 17.75\% & \textbf{18.70}\% \\ 
\hline\hline
\multirow{6}{*}{\rot{$\epsilon = 0.006$}} & js-AGCN & 8.23\% & ——      & 18.51\% & 18.88\% & 18.79\% & 20.74\% & 21.88\% & \textbf{22.06}\% \\
& 2s-AGCN & 5.09\% & ——      & 7.93\%  & 9.80\%  & 10.56\% & 9.80\%  & 11.26\% & \textbf{12.66}\% \\
& ST-GCN  & 9.54\% & 17.51\%  & 17.60\%  & 17.97\%  & 18.67\%  & 18.90\%  & 19.91\% & \textbf{20.34}\%  \\
& MS-G3D  & 9.42\% & 18.07\% & 15.06\% & 15.09\% & 15.20\% & 16.86\% & 17.25\% & \textbf{17.56}\% \\
& SGN     & 16.82\% & 34.10\% & 22.39\% & 22.65\% & 22.76\% & 25.43\% & 25.79\% & \textbf{25.92}\% \\
& ASGCN   & 5.39\% & 19.02\% & 12.90\%  & 13.45\% & 14.97\%  & 14.08\% & 15.26\% & \textbf{16.33}\% \\ 
\hline
\end{tabular}
\caption{The fooling rate of different methods on the target models in NTU120, where $\epsilon$ is the perturbation budget.}\vspace{-1em}
\label{120}
\end{table*}

\begin{table*}[htb]
\setlength\tabcolsep{4.5pt}
\scriptsize
\centering
\begin{tabular}{|c|c|cc|cccccc|}
\cline{2-10}
  \multicolumn{1}{c|}{} &
  \makecell{Victim\\models} &
  \makecell{Self-sup\\Attacker} &
  \makecell{AGCN\\Attacker} &
  \makecell{No-box\\I-FGSM} &
  \makecell{No-box \\$\rm S_{1}$I-FGSM} &
  \makecell{No-box \\$\rm S_{2}$I-FGSM} &
  \makecell{No-box\\MI-FGSM} &
  \makecell{No-box \\$\rm S_{1}$MI-FGSM} &
  \makecell{No-box \\$\rm S_{2}$MI-FGSM} \\ \hline
\multirow{6}{*}{\rot{$\epsilon = 0.01$}} & js-AGCN & 10.12\% & ——      & 10.61\% & 10.98\% & 11.55\% & 13.45\% & \textbf{14.20}\% & \textbf{14.20}\% \\
& 2s-AGCN & 1.89\%  & ——      & 5.30\% & 5.68\% & 5.68\% & 6.44\% & 6.44\% & \textbf{6.82}\% \\
& ST-GCN  & 6.44\%  & 24.26\% & 9.47\%  & 9.85\%  & 9.47\%  & 8.90\%  & 11.17\% & \textbf{11.74}\% \\
& MS-G3D  & 4.17\%  & 86.95\% & 41.67\% & 44.51\% & 43.94\% & 53.03\% & \textbf{56.06}\% & 53.41\% \\
& SGN     & 1.89\% & 3.98\%      & 2.46\% & 2.84\% & 3.03\% & 3.03\% & 3.40\% & \textbf{3.59}\% \\
& ASGCN   & 1.89\%      & 27.57\%      & 3.22\% & 4.36\% & 3.22\% & 4.36\% & \textbf{5.68}\% & \textbf{5.68}\% \\ 
\hline\hline
\multirow{6}{*}{\rot{$\epsilon = 0.008$}} & js-AGCN & 9.46\% & ——      & 8.14\% & 9.09\% & 9.09\% & 10.61\% & 10.98\% & \textbf{11.55}\% \\
& 2s-AGCN & 1.70\%  & ——      & 4.36\% & 4.92\% & 4.73\% & 4.55\% & 5.30\% & \textbf{5.87}\% \\
& ST-GCN  & 3.79\%  & 21.69\% & 8.33\%  & 8.52\%  & 8.52\%  & 8.33\%  & \textbf{8.90}\% & 8.33\%  \\
& MS-G3D  & 3.60\%  & 82.17\% & 25.76\% & 30.68\% & 29.17\% & 36.36\% & \textbf{39.39}\% & 39.02\% \\
& SGN     & 1.51\%      & 3.60\%      & 0.57\% & 1.51\% & 2.08\% & 2.27\% & 2.27\% & \textbf{2.46}\% \\
& ASGCN   & 1.70\% & 22.43\% & 2.46\%  & 2.84\% & 2.84\%  & 2.27\% & 2.84\% & \textbf{3.03}\% \\ 
\hline\hline
\multirow{6}{*}{\rot{$\epsilon = 0.006$}} & js-AGCN & 7.19\% & ——      & 5.11\% & 7.01\% & 6.44\% & 6.06\% & 6.63\% & \textbf{7.77}\% \\
& 2s-AGCN & 1.33\% & ——      & 2.08\%  & 2.08\%  & 2.27\% & 3.79\%  & 3.98\% & \textbf{4.17}\% \\
& ST-GCN  & 3.40\% & 19.85\%  & 6.44\%  & 7.01\%  & 6.63\%  & 7.58\%  & \textbf{8.33}\% & 7.77\%  \\
& MS-G3D  & 2.84\%  & 72.43\% & 11.55\% & 12.31\% & 11.92\% & 15.34\% & \textbf{18.94}\% & 17.99\% \\
& SGN     & 0.57\% & 3.40\% & 0.13\% & 0.94\% & 0.57\% & 1.13\% & 1.32\% & \textbf{1.51}\% \\
& ASGCN   & 1.52\% & 17.82\% & 0.38\%  & 0.76\% & 1.89\%  & 0.57\% & \textbf{2.27}\% & \textbf{2.27}\% \\ 
\hline
\end{tabular}

\caption{The fooling rate of different methods on the target models in HDM05, where $\epsilon$ is the perturbation budget.\vspace{-1em}
}
\label{2}
\end{table*}

\section{Experiments}
We refer the readers to the supplementary material for extra experimental results.

\textbf{Datasets} 
We select three widely used skeletal datasets: HDM05 \cite{muller2007mocap} (2,337 sequences of 130 classes performed by 5 actors), NTU60 \cite{RN35} (56,880 sequences of 60 classes), and NTU120 \cite{liu2019ntu} (114,480 sequences of 120 classes, an extended version of NTU60, one of the largest datasets in the field). We pre-process HDM05 following \cite{du2015hierarchical} and both NTU datasets following \cite{RN33}. The different skeletons are mapped to a standard 25-joint structure as in \cite{RN12}. For our hard no-box attack, we only use the testing data and do not use the training data, the training labels and the testing labels during attacks.

\textbf{Target Models}
We choose multiple state-of-the-art models as victims: ST-GCN \cite{RN28}, 2s-AGCN \cite{RN33}, AS-GCN \cite{RN38}, SGN \cite{RN37} and MS-G3D \cite{RN39}. They are trained using the official implementations and following the training protocols. For 2s-AGCN, we attack both the single joint stream model (js-AGCN) and the two-stream model. 

\textbf{Implementation Details}
We pre-train the CL encoder $f_q$ following \cite{RN15}. The unsupervised network is trained with a temperature value $\tau=0.07$ and SGD optimizer for 450 epochs. The learning rate is set to 0.01 with a weight decay of 0.0001. Due to the limitation of hard no-box settings, the attacker cannot access the training samples during the whole process. Therefore, the encoder is trained on the testing set. We adopt the $l_\infty$ norm for the perturbation budget $\epsilon$. For clusters of negative samples, the number of clusters is 120, and the number of deleted centers $Q$ is 10.

\textbf{Evaluation Metrics}
We employ the fooling rate as a major metric. It is defined as the percentage of data samples whose predicted labels changed after adversarial attacks \cite{RN10}. Besides, inspired by \cite{RN5}, we define a perceptual deviation indicator to evaluate the imperceptibility of adversarial skeletal samples:
\begin{equation}
\begin{aligned}
\Delta p=\frac{1}{MT} \sum_{n=0}^{M}\left\|S-\hat{S}\right\|_{2}+
\frac{1}{MT} \sum_{n=0}^{M}\left\|B-\hat{B}\right\|_2\\+\frac{1}{MTL}\sum_{n=0}^{M}\left\|\ddot{S}-\ddot{\hat{S}}\right\|_{2}
\label{perceptual}
\end{aligned}
\end{equation}
where $M$ is the number of adversarial samples, $T$ is the total number of frames, and $L$ is the number of joints in the skeleton. The three terms evaluate the deviations of joint position, bone-length, and acceleration, respectively. 
A smaller perceptual deviation indicates better imperceptibility.
\begin{figure*}[htb] 
\centering
\includegraphics[width=0.9\textwidth]{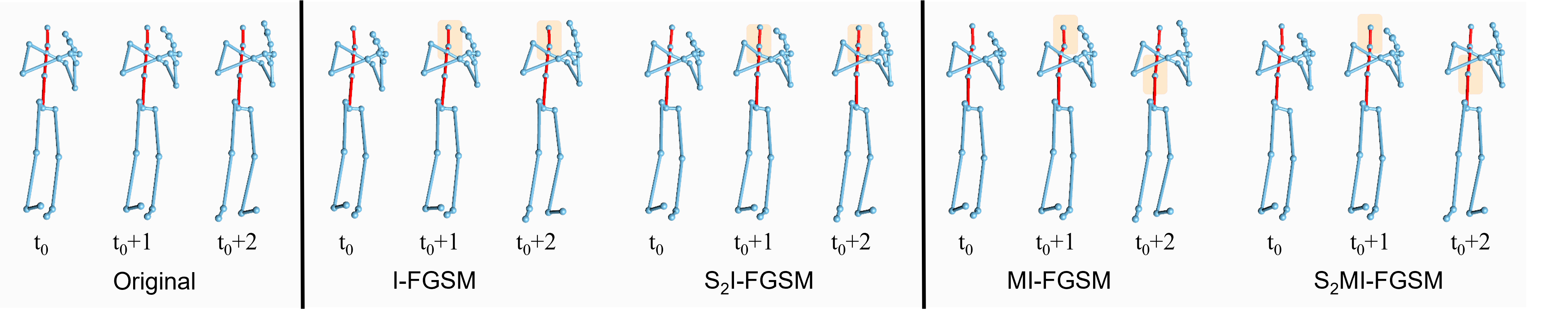} 
\caption{Visual comparisons between attack strategies in no-box attacks ($\epsilon$=0.006) with key visual differences highlighted.}\vspace{-1em} 
\label{Com} 
\end{figure*}
\subsection{Hard No-Box Attack}

\paragraph{Baselines} We establish two baselines using transfer-based attacks based on a self-supervised and a supervised classifier. This is because our method is the first hard no-box attack and there is no other similar method. The transfer-based attack is the closest setting to ours as they do not require access to the victim model. The first baseline (Self-sup Attacker) is a self-supervised surrogate model with a linear layer appended to our CL encoder. We freeze our CL encoder after the manifold training and then train the linear layer supervisedly \cite{RN27}. In this way, the need to access labels is only for training the linear layer, not the CL encoder, which is not strictly hard no-box but closer than existing methods. The second baseline is a standard transfer-based attack where we use js-AGCN as the surrogate model and SMART~\cite{RN5} as the white-box attacker (AGCN Attacker). Unlike our method, both baselines still require access to the training data and the training labels during attacks.

For hard no-box attack, different attack strategies are compared including I-FGSM, MI-FGSM, $\rm S_{1}$I-FGSM, $\rm S_{2}$I-FGSM, $\rm S_{1}$MI-FGSM, and $\rm S_{2}$MI-FGSM. We run all the attackers for 400 iterations and report the attack performance under different perturbation budgets $\epsilon$ in Table \ref{1} for NTU60, Table \ref{120} for NTU120, and Table \ref{2} for HDM05. We omit the results on js-AGCN and 2s-AGCN under AGCN Attacker because AGCN is the surrogate model.

Table \ref{1}-\ref{2} show that the hard no-box attack poses real threats to a range of S-HAR classifiers. In general, the fooling rate of hard no-box attacks is higher than Self-sup Attackers. This is surprising as the Attacker utilizes training data and training labels while our method does not. Looking further, AGCN Attacker is much better than Self-sup Attacker and the major difference is their feature extraction (i.e. one is a CL encoder and one is a graph network). This shows transfer-based attack heavily relies on the feature extraction ability of the surrogate model that cannot bypass access to the training data and the training labels. Furthermore, when compared with AGCN Attacker, our method achieves similar fooling rates, varying across different victims and datasets. Given that AGCN Attacker requires access to training data, training labels and testing labels, we argue hard no-box attacker achieves superior results and provides a more realistic setting.

In addition, among various no-box attack strategies, $\rm S_2$MI-FGSM performs the best and often by big margins. All the SMI gradient-based methods generate stronger adversaries compared with baselines I-FGSM and MI-FGSM. The 2nd-order SMI attack method usually outperforms the corresponding 1st-order version. Last, we notice a variance in fooling rate across different victims and datasets. For instance, the multi-stream model (2s-AGCN) significantly enhances the robustness compared to the single-stream model (js-AGCN); the fooling rate of all methods drops by nearly half. This may be because the multi-stream model can ensemble features from different modalities, which improves the robustness. In general, it is still an open question why fooling rate can vary across victims and datasets and we leave the theoretical analysis for future work.

\begin{table}[htb]
\setlength\tabcolsep{2pt}
\scriptsize
\centering
\begin{tabular}{|c|c|cccc|}
\hline
\makecell{Surrogate\\Model} & Victims & I-FGSM   & \makecell{MI-\\FGSM} & \makecell{$\rm S_2$I-FGSM\\(Ours)} & \makecell{$\rm S_2$MI-FGSM\\(Ours)}\\ \hline
\multirow{2}{*}{2s-AGCN} & STGCN        & 2.10\% & 2.10\%       & 3.00\%       & \textbf{3.01}\%        \\
& MS-G3D       & 2.58\% & 2.59\%       & 2.90\%       & \textbf{2.97}\%        \\ 
\hline\hline

\multirow{2}{*}{ST-GCN} & 2sAGCN        & 2.20\% & 2.34\%       & 2.44\%       & \textbf{2.64}\%        \\
& MS-G3D       & 2.00\% & 2.10\%       & 2.63\%       & \textbf{2.92}\%        \\ 
\hline\hline

\multirow{2}{*}{MS-G3D} & STGCN        & 1.71\% & 1.69\%       & 2.65\%       & \textbf{2.67}\%        \\
& 2sAGCN      & 1.76\% & 1.79\%       & 2.03\%       & \textbf{2.07}\%        \\ \hline
\end{tabular}
\caption{The fooling rate of different attack strategies in transferred SMART attacks, where attack budgets $\epsilon=0.01$.}\vspace{-1em}
\label{6}
\end{table}
\subsection{Transfer-Based Black-Box Attack} 
SMI gradient not only improves the transferability in hard no-box attacks, but also enhances other gradient-based skeletal attacks. Here, we employ SMART \cite{RN5}, a white-box attacker, as a baseline to compare different attack strategies. In the original SMART settings, I-FGSM is adopted to generate adversarial samples. We replace it with $\rm S_2$I-FGSM, MI-FGSM, and $\rm S_2$MI-FGSM to make a comparison. As all strategies achieve similar fooling rates in white-box attacks, we mainly focus on their transferability using different surrogate models. We utilize SMART to attack 2s-AGCN, ST-GCN, and MS-G3D on NTU60 and transfer the obtained samples to other victim networks.

The performance of the transferred black-box attack is shown in Table \ref{6}. $\rm S_2$MI-FGSM gives the best performance in transfer-based black-box attacks. $\rm S_2$I-FGSM also improves the transferability of adversarial samples compared with baselines. In contrast, MI-FGSM, which succeeds in the image transfer-based attack, struggles in skeletal data. Its performance declines to 1.69\% when it attacks STGCN via MS-G3D. Overall, the success rate is low in Table \ref{6} because SMART is sensitive to the chosen surrogate~\cite{RN5}. Transfer-based attacks are proven to suffer from lower fooling rates in S-HAR attack~\cite{RN5} and improving the transferability is still an open problem. Table \ref{6} aims to show our SMI gradient can improve it by incorporating motion dynamics into the attack gradient, compared with other alternative gradients. We will include in future work how to further explore this dynamics for better attack transfer.

\subsection{Perceptual Analysis}
A key feature of SMI gradient-based attacks is the improvement in the perceptual quality of the adversarial samples due to the consideration of the motion manifold. To verify this, we employ quantitative comparison and qualitative visual analysis on the no-box adversarial samples under various strategies. We compare the perceptual quality $\Delta p$ on NTU60 in Table \ref{3}. We find that $\rm S_2$I-FGSM achieves the best imperceptibility and obtains a massive improvement compared with I-FGSM. In contrast, MI-FGSM's deviation is twice that of $\rm S_2$I-FGSM. Although $\rm S_2$MI-FGSM does not achieve the best visual performance, it is still slightly better than I-FGSM and achieves a better trade-off between fooling rate and perceptual quality. This is understandable because our method considers dynamics that help to generate more on-manifold adversarial samples. Moreover, the 1st-order SMI attacks outperform baselines but cannot compete with 2nd-order SMI attacks. This demonstrates the importance of considering acceleration in the skeletal attack.

We show the visual comparison of poses under various attack strategies in no-box attacks in Figure \ref{Com}. The spinal joints demonstrate the most obvious differences. $\rm S_2$I-FGSM outperforms the other attack methods and gets the most natural poses, whereas I-FGSM has slight but noticeable joint displacements in the neck and head. $\rm S_2$MI-FGSM  performs better than its baseline, MI-FGSM, which shows zig-zag bending in the frame $t_0+2$. The samples produced by MI-FGSM have the worst imperceptibility, where we can easily find the unnatural jittery movements.

We also evaluate perceptual quality $\Delta p$ on SMART adversarial samples obtained with different gradients. Results conducted on the NTU60 dataset are shown in Table \ref{4}. $\rm S_2$I-FGSM reaches the best perceptual performance compared with all attack strategies. MI-FGSM slightly declines the imperceptibility. $\rm S_2$MI-FGSM performs slightly worse in MS-G3D and ST-GCN. The reason is mainly that $\rm S_2$MI-FGSM takes more iterations in the white-box attack, leading to late stopping and slightly worse visual performance.

\begin{table}[htb]
\centering
\scriptsize
\centering
\begin{tabular}{|c|ccc|}
\hline
Strategies & $\epsilon=0.01$         & $\epsilon=0.008$        & $\epsilon=0.006$         \\ \hline
I-FGSM             & 95.87          & 68.56          & 43.13           \\
MI-FGSM            & 131.77         & 90.44          & 54.61           \\
$\rm S_1$I-FGSM             & 84.63 & 60.77 & 38.15 \\
$\rm S_1$MI-FGSM            & 114.51          & 78.06          &   47.04       \\
$\rm S_2$I-FGSM             & \textbf{65.60} & \textbf{45.67} & \textbf{27.67 } \\
$\rm S_2$MI-FGSM            & 90.93          & 62.04          & 37.46  \\
\hline
\end{tabular}
\caption{The perceptual deviation of different attack strategies in no-box attacks with different budgets $\epsilon$.}\vspace{-1em}
\label{3}
\end{table}

\begin{table}[htb]
\setlength\tabcolsep{2pt}
\scriptsize
\centering
\begin{tabular}{|c|cccc|}
\hline
Victims & I-FGSM & MI-FGSM & \makecell{$\rm S_2$I-FGSM\\(Ours)} & \makecell{$\rm S_2$MI-FGSM\\(Ours)} \\ \hline
2s-AGCN       & 1.52  & 1.62    & \textbf{1.25}    & 1.49    \\
MS-G3D        & 2.39  & 2.46   & \textbf{1.69}   &  3.02  \\
ST-GCN        & 1.13  & 1.18   & \textbf{1.10}    & 1.47    \\\hline
\end{tabular}
\caption{The perceptual deviation of different attack strategies in SMART for different victim models.}\vspace{-1em}
\label{4}
\end{table}

\section{Conclusions and Discussions}
In this paper, we have verified potential threats to S-HAR solutions. A new setting is proposed: the hard no-box attack on skeletal motions without access to the victim model, the training samples or the labels. We validate our setting by proposing the first pipeline for hard no-box attacks. Moreover, as far as we know, we are the first to explore motion dynamics in the adversarial gradient computation, leading to a new SMI gradient compatible with existing gradient-based attacks. By extensive evaluation and comparison, our method has been proven to be threatening and imperceptible, relying on the least prior knowledge. 

The SMI gradient also improves the transferability of transferred black-box attacks. Nonetheless, boosting the transferability is still an open problem~\cite{RN5} and we will explore this further with our SMI gradient in the future. We will explore other models (e.g., diffusion models \cite{chang2023design}) and other time-series data (e.g. stock price, videos) for the proposed attack. Also, our SMI gradient describes dynamics and may be beneficial for motion synthesis \cite{chang2022unifying}. 

We call for attention to intensify the S-HAR robustness by considering defences against our hard no-box attack. We validate randomized smoothing \cite{cohen2019certified} as a potential defence method in supplementary materials. Due to the least prior knowledge requirement, security risks posed by our attack can be reduced with such defenses. Otherwise, our attacks become a significantly threat to S-HAR.

\section*{Acknowledegments}
This research is supported in part by National Natural Science Foundation of China (ref: 61673314, Yang), EPSRC (ref: EP/X031012/1, NortHFutures, Shum) and EU Horizon 2020 (ref: 899739, CrowdDNA, Wang).

{\small
\bibliographystyle{ieee_fullname}
\bibliography{egbib}
}
\clearpage
\includepdf[]{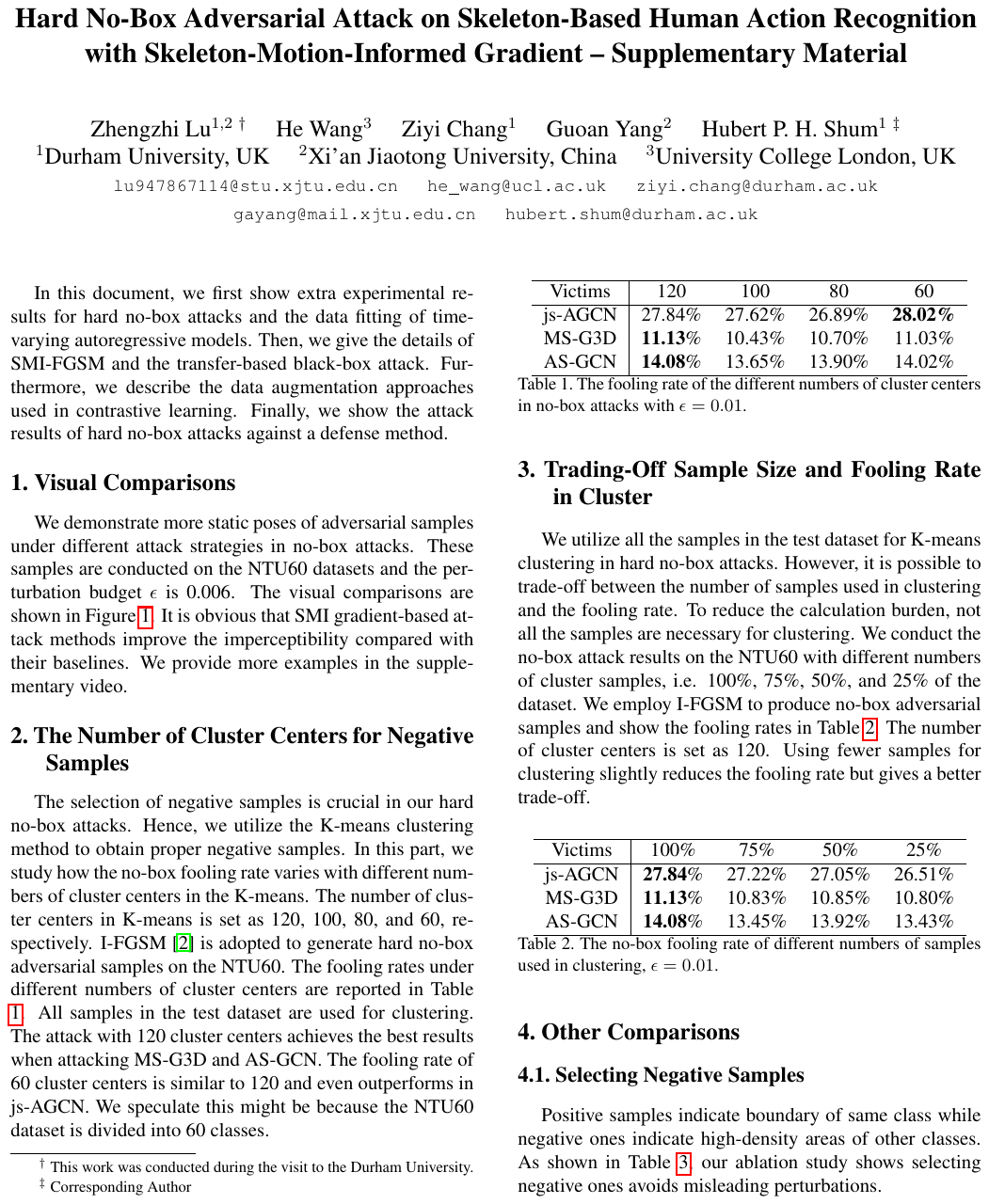}
\includepdf[]{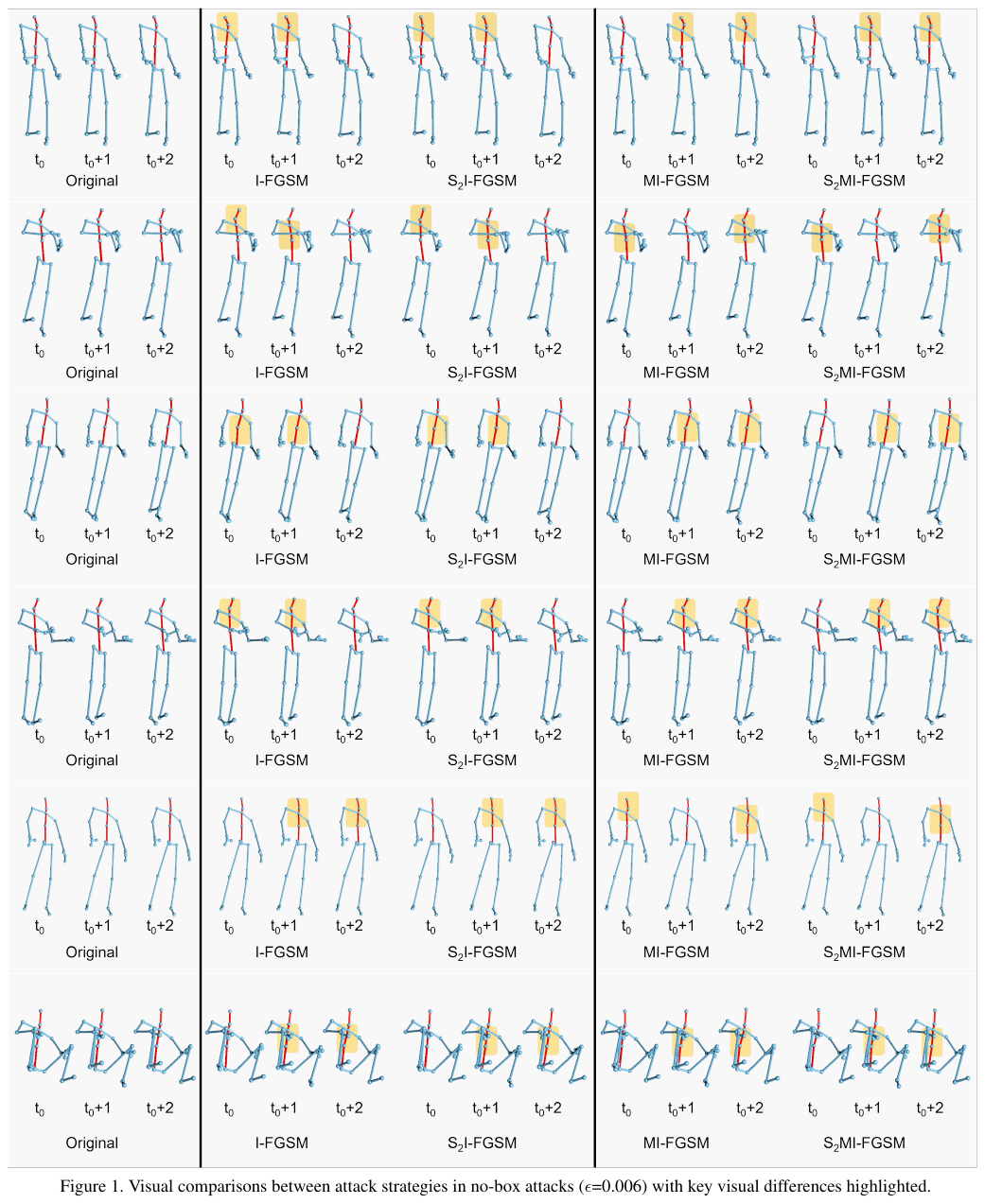}
\includepdf[]{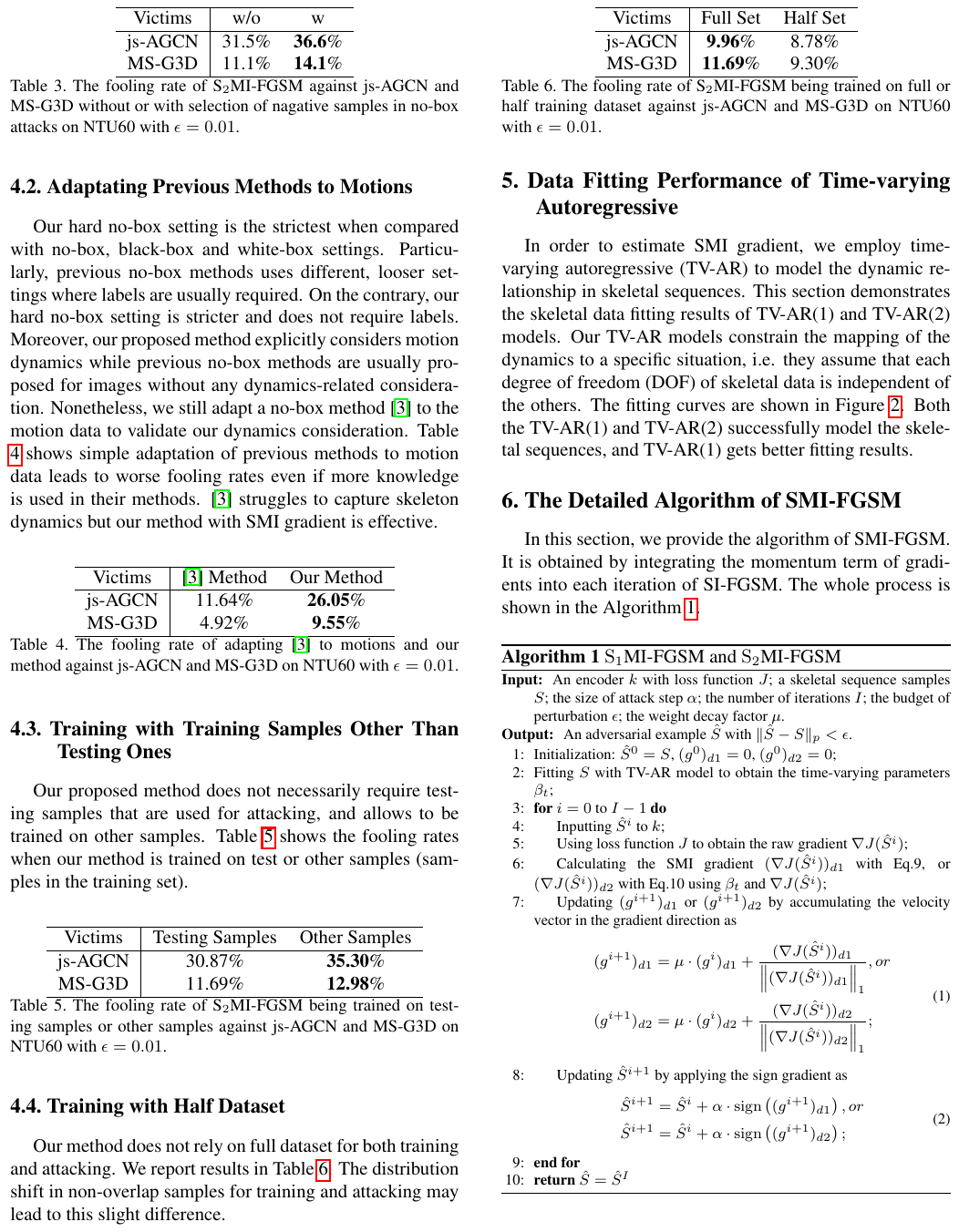}
\includepdf[]{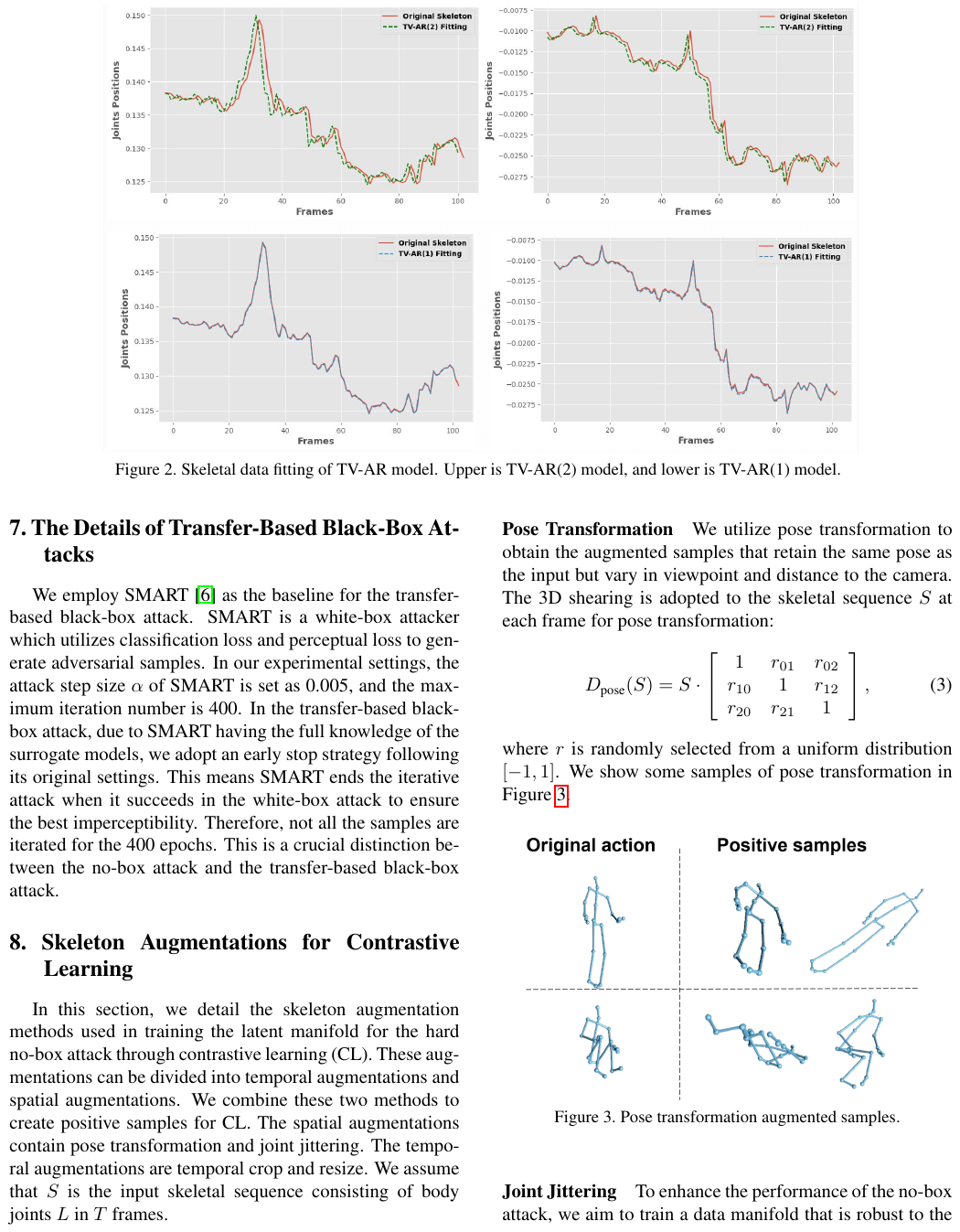}
\includepdf[]{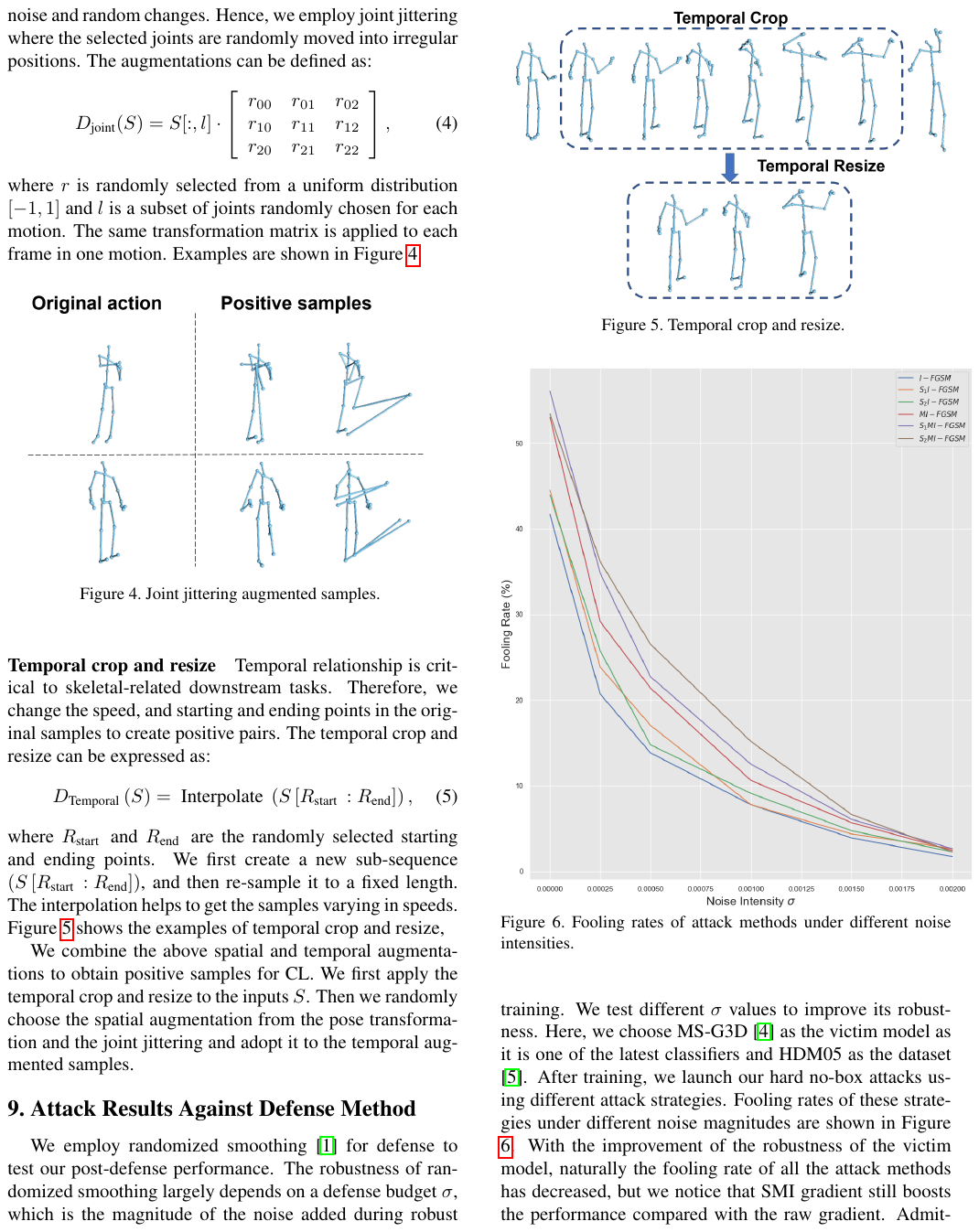}
\includepdf[]{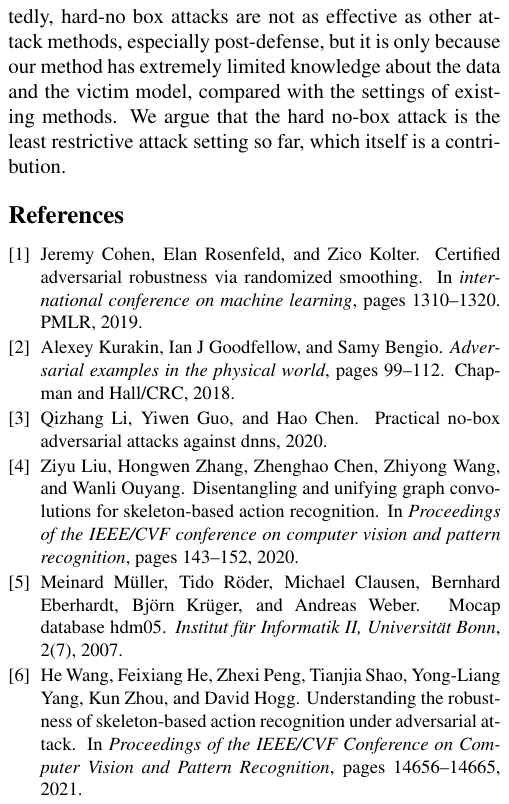}

\end{document}